\documentclass{paper}
\usepackage{booktabs} 
\usepackage[utf8]{inputenc}
\usepackage{url, hyperref}
\usepackage{cite, natbib}
\usepackage{graphicx}
\usepackage{amsmath, amsfonts, amsthm}

\newcommand{\attr}{\texttt}
\newtheorem{definition}{Definition}

\title{A comparative study of fairness-enhancing interventions in machine learning
\thanks{This work was partially supported by National Science Foundation
  under grants IIS-1633387, IIS-1513651, and IIS-1633724, as well as by a grant
  from the Ethics and Governance of AI Initiative.  Source code, including instructions for adding your own algorithm or dataset, can be found at: \url{https://github.com/algofairness/fairness-comparison}}
  }

\author{Sorelle A. Friedler, {\large Haverford College}\thanks{\url{sorelle@cs.haverford.edu}} \\
Carlos Scheidegger, {\large University of Arizona}\thanks{\url{cscheid@cscheid.net}} \\
Suresh Venkatasubramanian, {\large University of Utah}\thanks{\url{suresh@cs.utah.edu}} \\
Sonam Choudhary, {\large University of Utah}\thanks{\url{sonam@cs.utah.edu}} \\
Evan P. Hamilton, {\large Haverford College}\thanks{\url{evanphamilton@gmail.com}} \\
Derek Roth, {\large Haverford College}\thanks{\url{derek.roth17@gmail.com}}
}

\date{}

\begin{document}

\maketitle

\begin{abstract}
Computers are increasingly used to make decisions that have
significant impact in people's lives. Often, these predictions can
affect different population subgroups disproportionately. As a result,
the issue of \emph{fairness} has received much recent interest, and a
number of fairness-enhanced classifiers and predictors have appeared in
the literature. This paper seeks to study the following questions: how
do these different techniques fundamentally compare to one another,
and what accounts for the differences? Specifically, we seek to bring
attention to many under-appreciated aspects of such fairness-enhancing
interventions.  Concretely, we present the results of an open
benchmark we have developed that lets us compare a number of different
algorithms under a variety of fairness measures, and a large number of
existing datasets.  We find that although different algorithms tend to
prefer specific formulations of fairness preservations, many of these
measures strongly correlate with one another. In addition, we find
that fairness-preserving algorithms tend to be sensitive to
fluctuations in dataset composition (simulated in our benchmark by
varying training-test splits), indicating that fairness interventions
might be more brittle than previously thought.
\end{abstract}

\section{Introduction}
\label{sec:introduction}

As the use of machine learning to make decisions about people has increased, so has the drive to make fairness-aware machine learning algorithms.  A considerable body of research over the past ten years has produced algorithms for  accurate yet fair decisions, under varying definitions of fair, for goals such as non-discriminatory hiring, risk assessment for sentencing guidance, and loan allocation. And yet we have not yet seen extensive deployment of these algorithms in the pertinent domains. The primary obstacle appears to be our ability to compare methods effectively across different evaluation measures and different data sets with consistent data preprocessing and testing methodologies. Such comparisons would not just reveal ``best-in-class'' methods; they would also suggest which measures are robust and how different algorithms are sensitive to different kinds of preprocessing. As pointed out by \citet{lehr17:_playin_data}, such considerations of the data processing \emph{pipeline} are not just important for efficient implementation but also have legal ramifications for the resulting automated decision-making process. 

In this paper, we present a test-bed to facilitate direct comparisons of \emph{algorithms} with respect to \emph{measures} on a variety of \emph{datasets}. Our open-source framework allows for the easy addition of new methods, measures and data for the purpose of evaluation. We show how to use our test-bed for determining not only which specific algorithm has the best performance under a fairness or accuracy measure, but what types of algorithmic interventions tend to be the most effective.  In addition to the impact of these algorithmic choices, we examine the impact of different preprocessing techniques and different measures for accuracy and fairness that have an important, and previously obscured, impact on the results of these algorithms.  Our goal is to provide a comprehensive comparative analysis of existing approaches that is currently lacking in the literature.

\subsection{Our results}

In terms of the techniques, datasets, and measures we evaluate in this paper, we wish to highlight the following findings:

\paragraph*{Dependence on preprocessing} Different algorithms tend to have slightly different requirements in terms of input: how are sensitive attributes encoded? Are multiple sensitive attributes supported? Does the algorithm directly support categorical attributes or are attribute transformations required? We find that these can have an impact in accuracy and fairness measures reported in the literature.

\paragraph*{Clustering of measures} Even though there has been a proliferation of measures designed to highlight discrimination instances by machine learning algorithms, we find that a large number of these measures tend to strongly correlate with one another. As a result, techniques optimizing for one measure often performs well for a different measure (and similarly for poor performance).

\paragraph*{Algorithms make significantly different tradeoffs} The specific mechanisms that different algorithms employ to increase fairness are quite varied, but surprisingly, the actual predictions made by these algorithms tend to vary significantly as well. As a result, no algorithm's performance (as of the latest state of our benchmark)  appears to dominate, either in accuracy or fairness measures.

\paragraph*{Algorithms tend to be sensitive to variations in the input} We find surprising variability in fairness measures arising from variations in training-test splits; this appears to not have been previously mentioned in the literature.


\section{Background}
\label{sec:related-work}
Fairness-aware machine learning algorithms seek to provide methods under which
the predicted outcome of a classifier operating on data about people is fair or
non-discriminatory for people based on their \emph{protected class status} such
as race, sex, religion, etc., also known as a \emph{sensitive attribute}.
Broadly, fairness-aware machine learning algorithms have been categorized as
those \emph{preprocessing} techniques designed to modify the input data so that
the outcome of any machine learning algorithm applied to that data will be fair,
those \emph{algorithm modification} techniques that modify an existing algorithm
or create a new one that will be fair under any inputs, and those
\emph{postprocessing} techniques that take the output of any model and modify
that output to be fair  \cite{Romei13Multidisciplinary}.  Many associated
metrics for measuring fairness in algorithms have also been explored. These are
detailed further in Section \ref{sec:metrics} and are also surveyed in
\cite{Zliobaite2017Measuring}. This description of fairness-aware machine
learning methods is limited to batch-learning-based interventions. We do not consider interventions that focus on sequential or reinforcement learning such as \cite{jabbari_fairness_2017,joseph_fairness_2016,joseph_fair_2016,pm,pp}

\paragraph{Preprocessing algorithms} The motivation behind preprocessing algorithms is the idea that training data is the cause of the discrimination that a machine learning algorithm might learn, and so modifying it can keep a learning algorithm trained on it from discriminating.  This could be because the training data itself captures historical discrimination or because there are more subtle patterns in the data, such as an under-representation of a minority group, that makes errors on that group both more likely and less costly under certain accuracy measures.  One such algorithm that we will analyze in this paper is that of \citet{2015_kdd_disparate_impact} that modifies each attribute so that the marginal distributions based on the subsets of that attribute with a given sensitive value are all equal; it does not modify the training labels.  Additional preprocessing approaches include \cite{Flavio2017Preprocessing, KamiranCalders2012Preprocessing}.

\paragraph{Algorithm modifications}
Modifications to specific learning algorithms, e.g., in the form of additional
constraints, have been by far the most common
approach. We study three such methods in this paper.  \citet{Kamishima12Fairness} introduce a fairness focused
regularization term and apply it to a logistic regression
classifier. \citet{zafar2017fairness} observe that standard fairness constraints
are nonconvex and hard to satisfy directly and introduce a convex relaxation for
purpose of optimization. \citet{Calders10NaiveBayes} build separate models for
each value of a sensitive attribute and use the appropriate model for inputs
with the corresponding value of the attribute.

Another method that combines preprocessing and algorithm modification is the
work by \citet{icml2013_zemel13}. Their approach is to learn a modified
representation of the data that is most effective at classification while still
being free of signals pertaining to the sensitive attribute. 

\paragraph{Postprocessing techniques}
A third approach to building fairness into algorithm design is by modifying the
results of a previously trained classifier to achieve the desired results on
different groups. \citet{kamiran2010discrimination} designed a strategy to
modify the labels of leaves in a decision tree after training in order to
satisfy fairness constraints. Recent work by \citet{hardt2016equality} and
\citet{woodworth2017learning} explored the use of post-processing as a way to
ensure fairness with respect to error profiles (see Section~\ref{sec:metrics}
for more on this).

In this paper we focus on \emph{group fairness} approaches that aim to ensure
non-discrimination across protected groups where the goal is to optimize metrics
such as disparate impact.  Another line of thought, known as \emph{individual
  fairness}, is detailed in \cite{Dwork12Fairness}. In this work, we do not
study algorithms that seek to optimize individual fairness: our goal is to focus
on methods that explicitly deal with group-based discrimination and there are
(to the best of our knowledge) no actual codes that optimize for individual
fairness. 

\subsection{Related Work}
\label{sec:related-work}

Three prior efforts are relevant to our work. \textsc{FairTest}
\cite{DBLP:journals/corr/TramerAGHHHJL15}\footnote{\url{https://github.com/columbia/fairtest}}
provides a general methodology to explore potential biases or feature
associations in a data set, as well as a way to identify regions of the input
space where an algorithm might incur unusually high
errors. \textsc{THEMIS}\cite{galhotra2017fairness}\footnote{\url{https://github.com/LASER-UMASS/Themis}}
takes a blackbox decision-making procedure and designs test cases automatically to explore
where the procedure might be exhibiting group-based or causal discrimination. 
Fairness Measures
\cite{fairnessmeasures} occupies a different point in the design space. Given a
particular algorithm that one wishes to evaluate, they provide a framework to
test the algorithm on a variety of datasets and fairness measures. This approach
on the one hand is more general than our framework, because it works with any
algorithm. On the other hand, it is less effective for a comparative evaluation
of different algorithms especially if they have different preprocessing and
training methods.

There are other software packages that audit black box software to determine the
influence of individual variables. We omit a detailed description of these
approaches as they are out of the scope of the investigation presented here. For
more information, the reader is referred to the excellent new survey on
explainability by \citet{guidotti2018survey}. 


\section{Benchmark Structure}
\label{sec:setu}

\begin{figure*}[htb]
\begin{center}
\includegraphics[width=\textwidth]{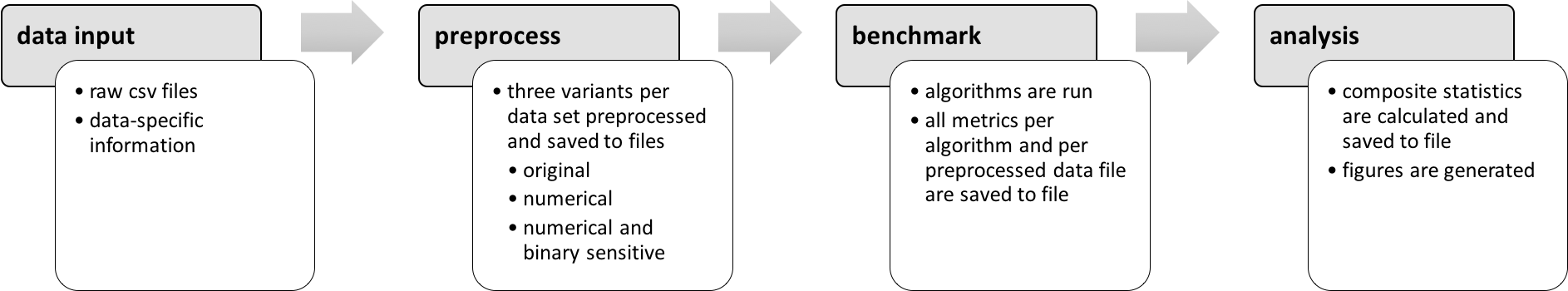}
\caption{The stages of the fairness-aware benchmarking program: data input, preprocessing, benchmarking, and analysis.  Intermediate files are saved at each stage of the pipeline to ensure reproducibility.}
\label{fig:flowchart}
\end{center}
\end{figure*}

In order to provide a platform for clear comparison of results across fairness-aware machine learning algorithms, we separate each stage of the learning and analysis process (see Figure \ref{fig:flowchart}) and ensure that each algorithm is compared using the same dataset (including the same preprocessing), the same set of training / test splits, and all desired fairness and accuracy measures.  Much previous work has combined the preprocessing for a specific dataset with the code for the fairness-aware algorithm, which makes comparisons with other algorithms and other datasets difficult.  Similarly, algorithms have often been analyzed only under one or two measures.  Here, we emphasize that we distinguish preprocessing, algorithms, and measures, and create a pipeline in which all algorithms are analyzed under a standard preprocessing of datasets and a large set of measures.

In order to encourage easy adoption of this codebase as a platform for future algorithmic analysis, each of these choices is modularized so that adding new datasets, measures, and/or algorithms to the pipeline is as easy as creating a new object.  The pipeline will then ensure that all existing algorithms are evaluated under the new dataset and measure.  More details and instructions for adding to the code base can be found at the repository.\footnote{\url{https://github.com/algofairness/fairness-comparison}}


\section{Data}
\label{sec:data}

We perform all experiments based on five real-world data sets that have been previously considered in the fairness-aware machine learning literature and preprocess each consistently depending on the needs of the algorithm.\footnote{All raw datasets, preprocessing code, and resulting processed datasets are available in the repository: \url{https://github.com/algofairness/fairness-comparison}.  Preprocessing described here can be reproduced by running: \texttt{python3 preprocess.py}}  The real-world datasets come from some of the domains impacted by questions of fairness in machine learning: hiring and promotion, credit-worthiness and loans, and recidivism prediction.

\paragraph{Ricci}  The Ricci dataset comes from the case of Ricci v. DeStefano \cite{Ricci09}, a case before the U.S. Supreme Court in which the question at issue was an exam given to determine if firefighters would receive a promotion.  The dataset has 118 entries and five attributes, including the sensitive attribute \attr{Race}.  The original promotion decision was made by a threshold of achieving at least a score of $70$ on the combined exam outcome \cite{Miao11RicciStats}.  The goal in a fair learning context is to predict this original promotion decision while achieving fairness with respect to the sensitive attribute, \attr{Race}. 

\paragraph{Adult Income}  The Adult Income dataset \cite{UCIrepo} contains information about individuals from the 1994 U.S. census.  It is pre-split into a training and test set; we use only the training data and re-split it.  There are 32,561 instances and 14 attributes, including sensitive attributes \attr{race} and \attr{sex}.  2,399 instances with missing data are removed during the preprocessing step.  The prediction task is predicting whether an individual makes more or less than \$50,000 per year.

\paragraph{German} The German Credit dataset \cite{UCIrepo} contains 1,000 instances and 20 attributes describing individuals along with a classification of each individual as a good or bad credit risk.  Sensitive attribute \attr{sex} is not directly included in the data, but can be derived from the given information.  Sensitive attribute \attr{age} is included, and is discretized into values \attr{adult} (age at least 25 years old) and \attr{youth} based on an analysis by \cite{Kamiran09Classifying} showing this discretization provided for the most discriminatory possibilities. 

\paragraph{ProPublica recidivism} The ProPublica data includes data collected about the use of the COMPAS risk assessment tool in Broward County, Florida \cite{propublica}.  It includes information such as the number of juvenile felonies and the charge degree of the current arrest for 6,167 individuals, along with sensitive attributes \attr{race} and \attr{sex}.  Data is preprocessed according to the filters given in the original analysis \cite{propublica}.  Each individual has a binary ``recidivism" outcome, that is the prediction task, indicating whether they were rearrested within two years after the first arrest (the charge described in the data).

\paragraph{ProPublica violent recidivism} The violent recidivism version of the ProPublica data \cite{propublica} describes the same scenario as the recidivism data described above, but where the predicted outcome is a rearrest for a violent crime within two years.  4,010 individuals are included after preprocessing is applied, and the sensitive attributes are \attr{race} and \attr{sex}.

\section{Preprocessing}
\label{sec:preprocessing}
Each algorithm we will analyze has certain requirements for the type of data it will operate over, and these necessitate different preprocessing techniques.  However, in order to provide a consistent comparison across algorithms, it's important that each algorithm receive the same input.  We reconcile these needs by creating types of inputs that multiple algorithms can handle.  Algorithms that handle the same input can then be easily compared to each other;  algorithms can also be compared across different preprocessing strategies for the same dataset, though these results should be seen to be less definitive.

The first preprocessing step is to modify the input data according to any data-specific needs: removing features that should not be used for classification, removing or imputing any missing data, and potentially removing items or adding derived features.  In order to allow the analysis of fairness based on multiple sensitive attributes (e.g., not just ensuring fairness based on race or sex alone, but based on both someone's race and sex) we also add a combined sensitive attribute (e.g., attribute ``race-sex" with values like ``White-Woman") to each dataset that contains multiple sensitive attributes.   All algorithms will receive versions of the dataset with this same preprocessing applied.

While some algorithms are able to handle the datasets for training with only the described initial preprocessing (we'll call this version of the processed data ``\texttt{original}"), most algorithms considered here have additional constraints.\footnote{Since scikit-learn classifiers only handle numerical data, even for classifiers like decision trees where this is not inherently a requirement, some of the tested algorithms that would otherwise handle the original data require numerical data since the algorithms call scikit-learn.}   For algorithms that can only handle \texttt{numerical} training data as input, we modify the data to include one-hot encoded versions of each categorical variable.  Some algorithms additionally require that the sensitive attributes be binary (e.g., ``White" and ``not White" instead of handling multiple racial categorizations) - for this version of the data (\texttt{numerical+binary}) we modify the given privileged group to be $1$ and all other values to be $0$.  

\subsection{Analysis}
\label{sec:analys-diff-prepr}

With these four preprocessed versions of each data set in place, we can compare how a single algorithm performs relative to all versions of the dataset on which it can run.  The most common form of input for the algorithms we consider here is \texttt{numerical}, and all these algorithms can additionally handle the \texttt{numerical+binary} version of the dataset.  This gives an opportunity to determine the effect, per algorithm and per dataset, of allowing an algorithm access to full information about sensitive attribute categorization or only a binary summary.

\begin{figure*}
  \centering
  \includegraphics[width=.45\linewidth]{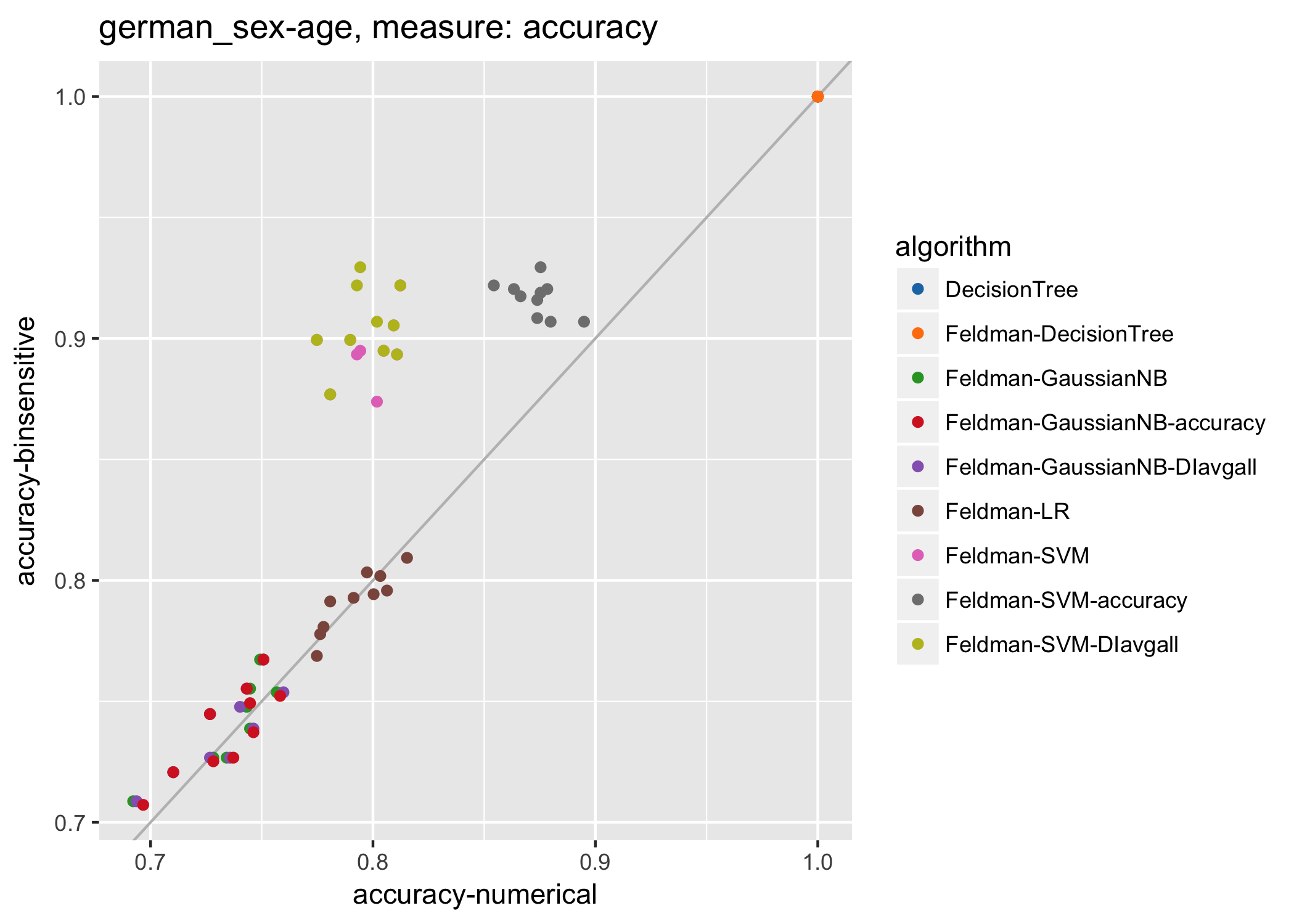}
  \includegraphics[width=.45\linewidth]{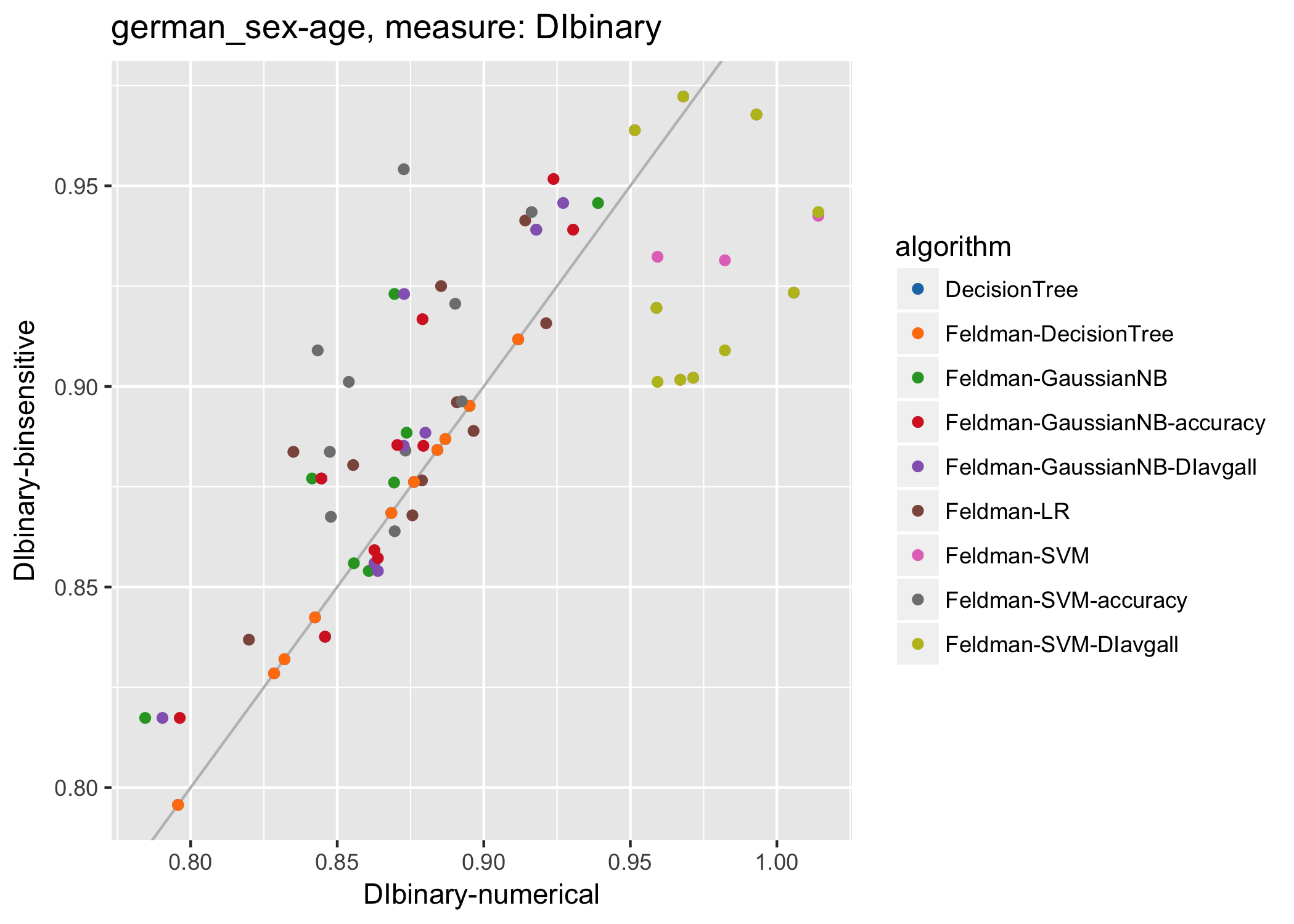}
  \caption{\label{fig:preprocessing-tradeoffs} Examining the results of the Feldman et al. \cite{2015_kdd_disparate_impact} algorithm under different preprocessing choices: \texttt{numerical} versus \texttt{numerical+binary}.  Each dot plots the result of a single split of the data in terms of the labeled metric under both preprocessing choices.  The gray line shows equality between the preprocessing choices.  The model used within the Feldman algorithm is listed, and some variants of the algorithm had the tradeoff parameter optimized for either accuracy or disparate impact value.}
\end{figure*}

Figure~\ref{fig:preprocessing-tradeoffs} illustrates this analysis on the impact of the \texttt{numerical+ binary} version of the preprocessed data on the Feldman et al. \cite{2015_kdd_disparate_impact} algorithm.  In the left figure we examine the relation between the accuracy on \texttt{numerical} preprocessing versus \texttt{numerical+binary} binary-encoded sensitive attributes. Each algorithm was run over ten random $\frac{2}{3}:\frac{1}{3}$ splits and the result on each split is shown as a single point on the figure. As discussed in Section~\ref{sec:algorithms}, \citeauthor{2015_kdd_disparate_impact} use a generic classifier after running a preprocessing ``fairness-enhancing'' filter on the data, and the different algorithms reflect the different classifier used.  We also automate the parameter tuning for $\lambda$, the fairness-accuracy tradeoff parameter for this algorithm (more about parameter tuning specifics can be found in Section \ref{sec:algorithms}), for both accuracy and the disparate impact value.  As we can see, for most variants of the algorithm the resulting accuracy is independent of the representation, with a notable exception of the SVM variants (where the preprocessing is followed by training with an SVM).  In all three SVM variants, the accuracy is consistently higher when using the \texttt{numerical+binary} representation than when using the \texttt{numerical} representation.  We speculate that this is because the Feldman et al. algorithm conditions on the sensitive value in its preprocessing on the data, and this step likely preserves more accuracy when a larger number of people are in each sensitive group -- as is the case when the unprivileged groups are grouped together in the binary preprocessing variant.  This may be compounded by the SVM model because when categorical features are one-hot encoded for input  (as required by \texttt{scikit-learn}) the increase in the dimensionality of the data may cause the SVM to be less effective at finding a good classifier.

We can do a similar analysis on the fairness achieved by the methods, as seen in the right side of Figure~\ref{fig:preprocessing-tradeoffs}. Again, we compare the fairness measure (in this case DI -- see Section~\ref{sec:metrics}) achieved for different data representations.  First, we see that the fairness achieved varies across runs, an issue we will return to when we discus measure stability. Second, we notice that there is less difference between the results obtained for different representations (although SVMs still show sensitivity to the representation). In other words, for this algorithm the accuracy is affected by the choice of classifier and representation, but not the fairness achieved.


\section{Measures}
\label{sec:metrics}

There are many ways to evaluate the accuracy and fairness of a model. Rather than be exhaustive,\footnote{An upcoming tutorial puts the number of fairness measures at 21  \cite{arvind2018tutorial}!} we will focus on representative measures for each aspect.  Let $D = (\mathbb{X}, S, Y)$ be a dataset where $\mathbb{X}$ is the data subset that can be used for training (whether categorical or numerical), $S$ is the sensitive attribute where $1$ is the privileged class, and $Y$ is the binary classification label where $1$ is the positive outcome and $0$ is the negative outcome.  Let $\hat{Y}$ be the predicted outcomes of some algorithm. We can define accuracy and fairness measures in terms of conditional probabilities of outcome variables ($Y, \hat{Y}$) with respect to variables like $\hat{Y}$ and $S$.

\subsection{Accuracy measures}
We consider the standard accuracy measures: the (uniform) accuracy ($P[\hat{Y} = y ~|~ Y = y] $), the true positive rate (TPR) ($P[\hat{Y} = 1 ~|~ Y = 1]$) (also called the positive predictive value (PPV)), and the true negative rate (TNR) ($P[\hat{Y} = 0 ~|~ Y = 0]$) (also called the negative predictive value (NPV)).  We also consider the balanced classification rate (BCR), a version of accuracy that is unweighted per class:

\begin{definition}[BCR]
\[ \frac{ P[\hat{Y} = 1 ~|~ Y = 1] + P[\hat{Y} = 0 ~|~ Y = 0] }{2}  \]
\end{definition}

All of these measures lie in the range $[0,1]$.

\subsection{Fairness measures}

Fairness measures can be divided into three broad categories, in all cases conditioned on values of the sensitive attribute $S$. In what follows, we normalize measures to make comparisons easier. In all cases, the measures lie in the range $[0, \infty)$ or $[0,2]$ where in both cases perfect fairness is achieved at $1$. We note that some of these measures have appeared in the literature not as something to be optimized (to be close to $1$) but as a constraint to be satisfied (i.e for example that the appropriate value must equal $1$). 

\subsubsection{Measures based on base rates}

\begin{definition}[Disparate Impact (DI) \cite{2015_kdd_disparate_impact,zafar2017fairness}]
\[ \frac{ P[\hat{Y} = 1 ~|~ S \not= 1] }{ P[\hat{Y} = 1 ~|~ S = 1]} \]
\end{definition}

This measure is inspired by one of the two tests for disparate impact in the legal literature in the United States\cite{barocas2016big}. 
In the cases where there are more than two values for a given sensitive attribute, we consider two variants of DI (which are equivalent in the case when there are only two sensitive values): binary and average.  In the binary case, all unprivileged classes are grouped together into a single value $S \ne 1$ (e.g., "non White") that is compared as a group to the privileged class $S=1$ (e.g., "White").  In the average case, pairwise DI calculations are done against the privileged class (e.g., "White" compared to "Black", "White" compared to "Asian", etc.) and the average of these calculations is taken. This is analogous to the one-vs-all and all-vs-all methodology in multi-class classification. 

\begin{definition}[CV \cite{Calders10NaiveBayes}]
\[ 1 - \left( P[\hat{Y} = 1 ~|~ S = 1] - P[\hat{Y} = 1 ~|~ S \not= 1] \right) \] 
\end{definition}

This measure is the same as DI, but where the difference is taken instead of the ratio; such a measure has been used for example to measure gender discrimination in the United Kingdom.  A binary grouping strategy (described above for DI) is used in the case where there is more than one sensitive value, and the averaging method can also be used. Note that we do not take the absolute value of the difference so that skew in favor of one group versus another can be detected. We note that  requiring $CV=1$ is called the \emph{demographic parity} constraint in the literature. 

\subsubsection{Measures based on group-conditioned accuracy}
\label{sec:measures-based-class}

In general, we can think of fairness measures based on group-conditioned accuracy as asking whether the error rates for each group are similar. This yields the following definitions. 

\begin{definition}
(\emph{Group-conditioned fairness measures.})
  \begin{description}
\item[$s$-Accuracy.] \[ P[\hat{Y} = y \mid Y = y, S = s] \]
\item[$s$-TPR.] \[ P[\hat{Y} = 1 \mid Y = 1, S = s]\]
\item[$s$-TNR.] \[ P[\hat{Y} = 0 \mid Y = 0, S = s]\]
\item[$s$-BCR.] \[\frac{ P[\hat{Y} = 1 \mid Y = 1, S = s] + P[\hat{Y} = 0 \mid Y = 0, S = s] }{2}\]
\end{description}
\end{definition}

We note that these measures have been studied under different names. For example, error rate balance \cite{chouldechova2017fair} is the aim of achieving equal $1 -$ $s$-TPR and $1-$ $s$-TNR values across sensitive groups.  Equalized odds \cite{hardt2016equality} is the aim of achieving equal $s$-TPR and $1 - $ $s$-TNR (the \emph{false positive rate}) across sensitive groups.

Letting any of the above measures be denoted $f(Y, \hat{Y}, s)$, the values can then be aggregated for comparison by taking the mean directly $\sum_{s \in S} f(Y, \hat{Y}, s) / |S|$ or by taking the mean over comparisons analogous to DI and CV: 
$f(Y, \hat{Y},s) / f(Y, \hat{Y},1)$ or $1 - (f(Y, \hat{Y},1) - f(Y, \hat{Y},s))$.  In each of these cases, as we saw  above, the unprivileged sensitive values could be grouped together or handled separately in the ratio or difference.

\subsubsection{Measures based on group-conditioned calibration}
\label{sec:measures-based-group}

A predictor that outputs a probability $\hat{Y}$ for an event is said to be \emph{well-calibrated} if $P[ Y = 1\mid \hat{Y} = p] = p$. Motivated by this, we can define fairness measures by group conditioning the calibration function. 

\begin{definition}[$s$-Calibration+]
\[ P[ Y = 1 \mid \hat{Y} = 1, S = s]  \]
\end{definition}

\begin{definition}[$s$-Calibration-]
\[ P[ Y = 1 \mid \hat{Y} = 0, S = s]  \]
\end{definition}

Calibration has been introduced previously with the goal of equalizing across sensitive value \cite{chouldechova2017fair, kleinberg2017inherent}.

\subsection{Analysis}
\begin{figure}
  \hspace{-3em}\includegraphics[width=1.05\linewidth]{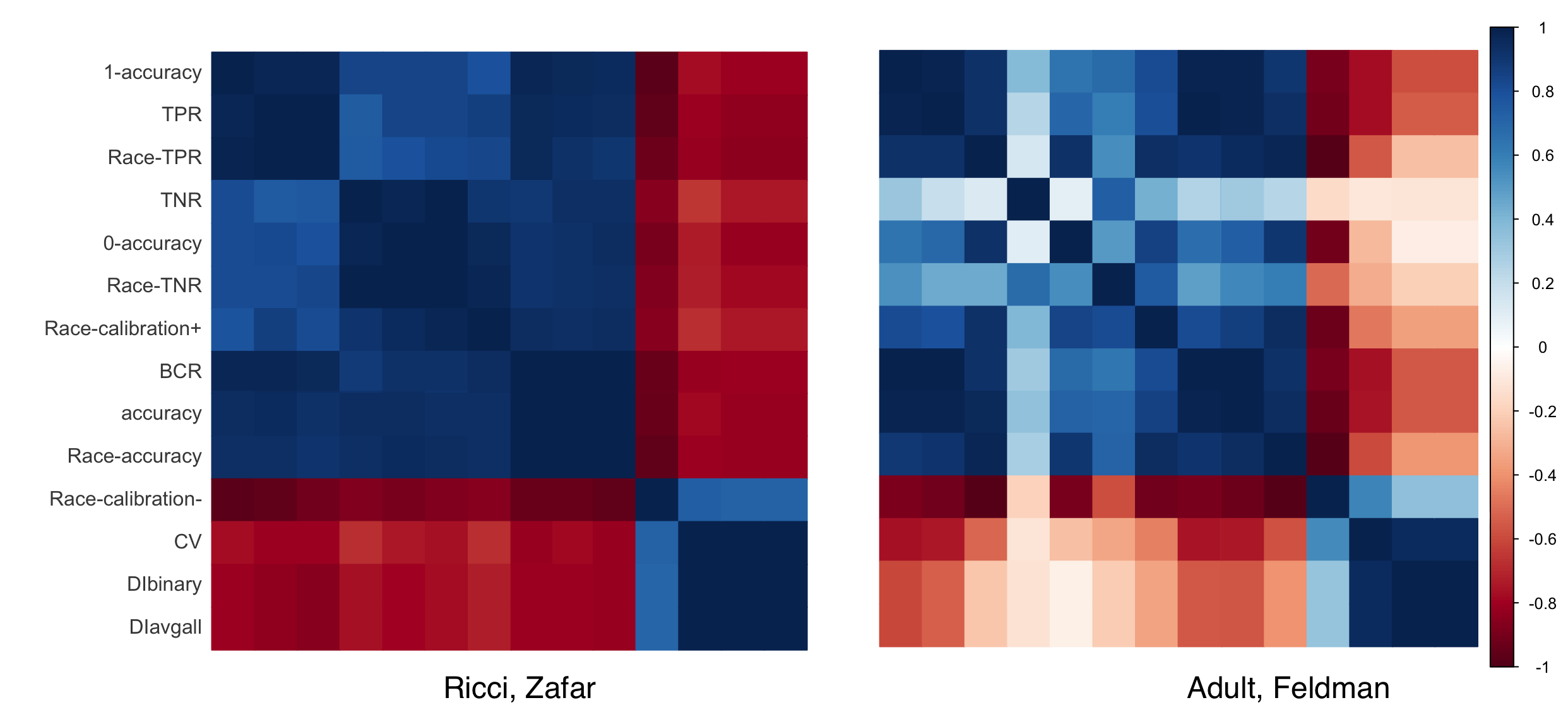}
  \caption{Examining the relationships between different measures of fairness. Each figure represents one data set-algorithm pair. For each entry, the algorithm is run for 10 training-testing splits for different parameter choices. The Stahel-Donoho estimator\cite{stahel1981breakdown,donoho1982breakdown} is then computed for each set of pairs of measurements.\label{fig:measure-correlation}}
\end{figure}

Although there are many variations on these and other measures, we find that many of these are correlated.  In some cases, this is not surprising as these measures are definitionally related.  For example, DI takes the ratio of two probabilities while CV takes the difference.  However, by analyzing resulting measures across many algorithms, we can find correlations that are less obvious.  In fact, it appears that there are two main groups of measures, all correlated with each other!  In Figure \ref{fig:measure-correlation} we fix two dataset-algorithm pairs and look at how the different measures of fairness correlate with each other. A first surprising observation is that the various group-conditioned fairness measures are very closely related to each other (the base-rate measures like DI and CV are also closely related for the reason mentioned above). This suggests that we need not focus on the specific group-conditioned fairness measure we use. An unusual exception to this is the group-conditional calibration measure on negative outcomes (\texttt{s-Calibration-}) which is much more closely associated with the base-rate measures than other group-conditioned measures. A second surprising observation is that the accuracy measures are correlated with the group-conditioned fairness measures. This suggests that the discussions of fairness-accuracy tradeoffs are more pertinent with respect to base-rate fairness measures. 

\begin{figure}[ht]
\begin{center}
\includegraphics[width=\textwidth]{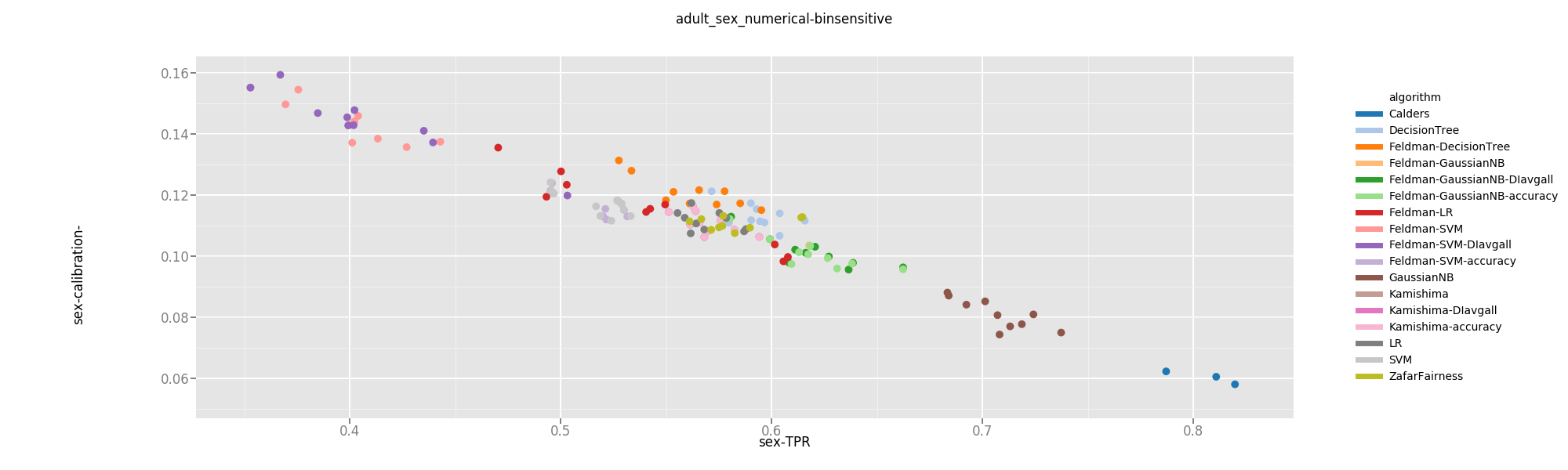}\\
\caption{An illustration of the tradeoff between \texttt{s-calibration-} and \texttt{TPR} for all algorithms on the \texttt{Adult} dataset. Each dot represents one run out of 10 random train-test splits.} 
\label{fig:tradeoff}
\end{center}
\end{figure}

Additionally, there are cases in which we would expect there to be tradeoffs between measures.  Recent impossibility results show that, assuming unequal base rates across populations, it is impossible to achieve both calibration and error rate balance (both the same false positive rate and the same false negative rates across groups) \cite{chouldechova2017fair, kleinberg2017inherent}.  In Figure \ref{fig:tradeoff} we empirically examine this tradeoff. As before, each colored point represents one instance of train-test split for an algorithm. As Figure~\ref{fig:tradeoff} shows, there is a clear tradeoff between with \texttt{s-calibration-} versus \texttt{s-TPR} for each algorithm. Interestingly, different algorithms situate themselves in different parts of the tradeoff line. 

\section{Algorithms}
\label{sec:algorithms}
We choose a selection of existing fairness-aware algorithms to assess; these are chosen based on availability of source code and with the goal of choosing varying types of fairness interventions (e.g., preprocessing versus algorithm modification).  Each algorithm is run on each dataset and each metric is calculated on the predicted results.\footnote{All algorithm implementations can be found in the repository (\url{https://github.com/algofairness/fairness-comparison}), along with all resulting metric calculations, (see the \texttt{results/} directory).  The full set of results can be reproduced by running: \texttt{python3 benchmark.py}}  Synthesis statistics (such as stability) are then calculated and comparison graphs are produced.\footnote{Algorithm analysis code can be found in the repository (\url{https://github.com/algofairness/fairness-comparison}) and can be reproduced by running: \texttt{python3 analysis.py}}  We analyze the following algorithms along with non-fairness-aware algorithms chosen for a baseline comparison: SVM, decision trees, Gaussian naive Bayes, and logistic regression.

\paragraph{\citet{Calders10NaiveBayes}}
\citeauthor{Calders10NaiveBayes}  introduce a fairness-aware algorithm modification called Two Naive Bayes. Their approach trains separate models for the values and iteratively assesses the fairness of the combined model under the CV measure, makes small changes to the observed probabilities in the direction of reducing the measure, and retrains their two models.  This algorithm can handle both categorical and numerical input data, but requires that the given sensitive attribute be binary.  We use the \citet{Kamishima12Fairness} implementation of this algorithm.\footnote{\url{https://github.com/tkamishima/kamfadm/releases/tag/2012ecmlpkdd}} The algorithm has a $\beta$ prior parameter, which we search from $0$ to $1$ in increments of $0.1$.

\paragraph{\citet{2015_kdd_disparate_impact}}
\citeauthor{2015_kdd_disparate_impact} give a preprocessing approach that modifies each attribute so that the marginal distributions based on the subsets of that attribute with a given sensitive value are all equal; it does not modify the training labels.  Any algorithm can then be trained on the resulting ``repaired" data.  The algorithm can handle both categorical and numerical input data, but since we train scikit-learn classifiers based on this preprocessed data, our implementation can only handle numerical input.  Both binary and non-binary sensitive attributes can be handled.  A tuning parameter $\lambda$ is provided to tradeoff between fairness and accuracy, where $\lambda = 0$ gives the fairness of a regular non-fairness aware classifier and $\lambda = 1$ maximizes fairness.  $\lambda = 1$ is used as the default, and all values of $\lambda$ at increments of $0.05$ in $[0,1]$ are included when the algorithm is optimized using a grid search over the parameters.  The implementation comes from Feldman et al.  \cite{2015_kdd_disparate_impact} and \cite{adler2018auditing}.\footnote{\url{https://github.com/algofairness/BlackBoxAuditing}}

\paragraph{\citet{Kamishima12Fairness}}
\citeauthor{Kamishima12Fairness} introduce a fairness-focused regularization term and apply it to a logistic regression classifier.  Their approach requires numerical input and a binary sensitive attribute.  A tuning parameter $\eta$ is provided to tradeoff between fairness and accuracy, where $\eta = 1$ is the default.  When optimizing the parameter we use values between $0$ and $300$, with a finer grid used for the lower values of that range; these parameter choices are based on the experimental exploration of this parameter given in \cite{Kamishima12Fairness}.  We use the \citeauthor{Kamishima12Fairness} implementation of this algorithm.\footnote{\url{https://github.com/tkamishima/kamfadm/releases/tag/2012ecmlpkdd}}

\paragraph{ \citet{zafar2017fairness}}
\citeauthor{zafar2017fairness} re-express fairness constraints (which can be nonconvex) via a convex relaxation. This allows them to maximize accuracy subject to fairness and also maximize fairness subject to fairness constraints. They use two parameters: $c$ is a parameter that controls the degree of independence of the outcome and the sensitive attribute via a covariance calculation: setting $c = 0$ forces complete independence (and therefore fairness). The second parameter $\gamma$ fixes the degree of approximation they are willing to tolerate: the algorithm is only required to find an answer that is within a $1+\gamma$ factor of the optimal solution. In their experiments they set $\gamma = 0.5$ and vary $c$ as a linear  function of the corresponding covariance estimate for an unconstrained classifier.  When optimizing, we use values between $0.001$ and $1$ in 10 logarithmic steps.

\begin{figure}[htbp]
\begin{center}
  \includegraphics[width=\textwidth]{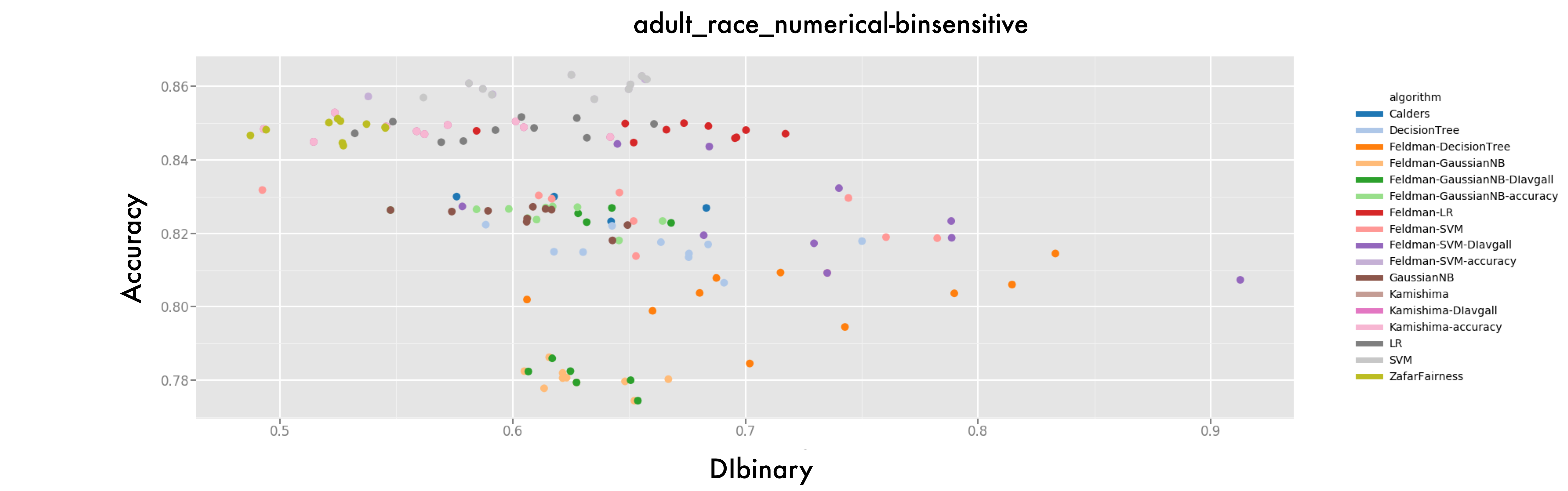}\\
  \includegraphics[width=\textwidth]{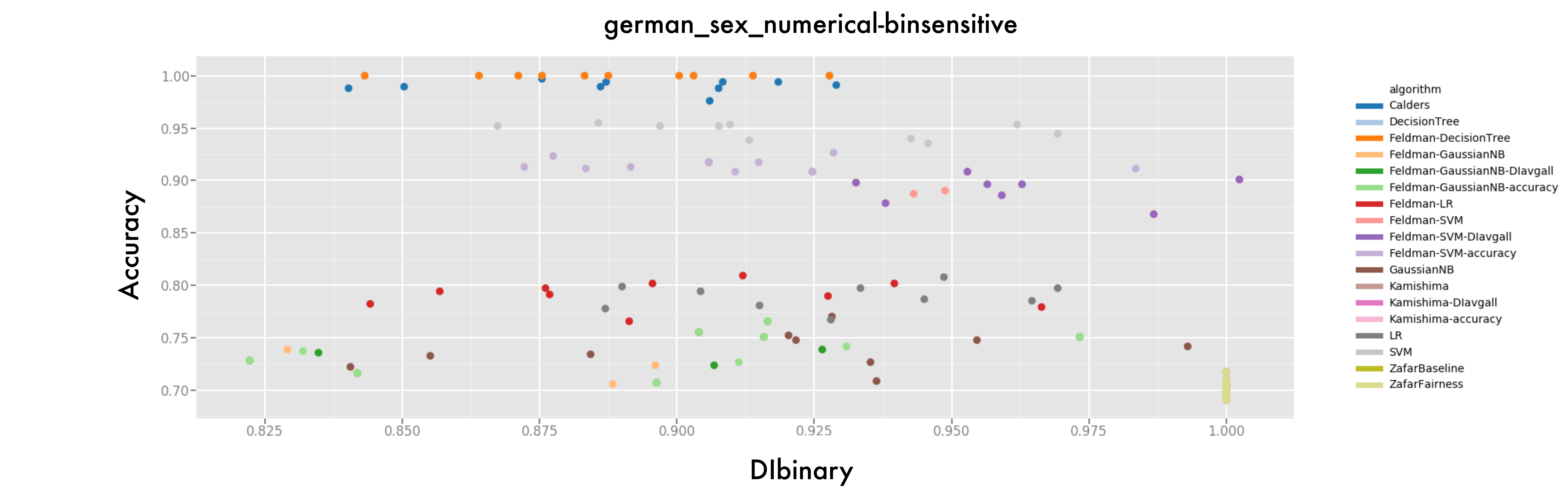}\\
  \includegraphics[width=\textwidth]{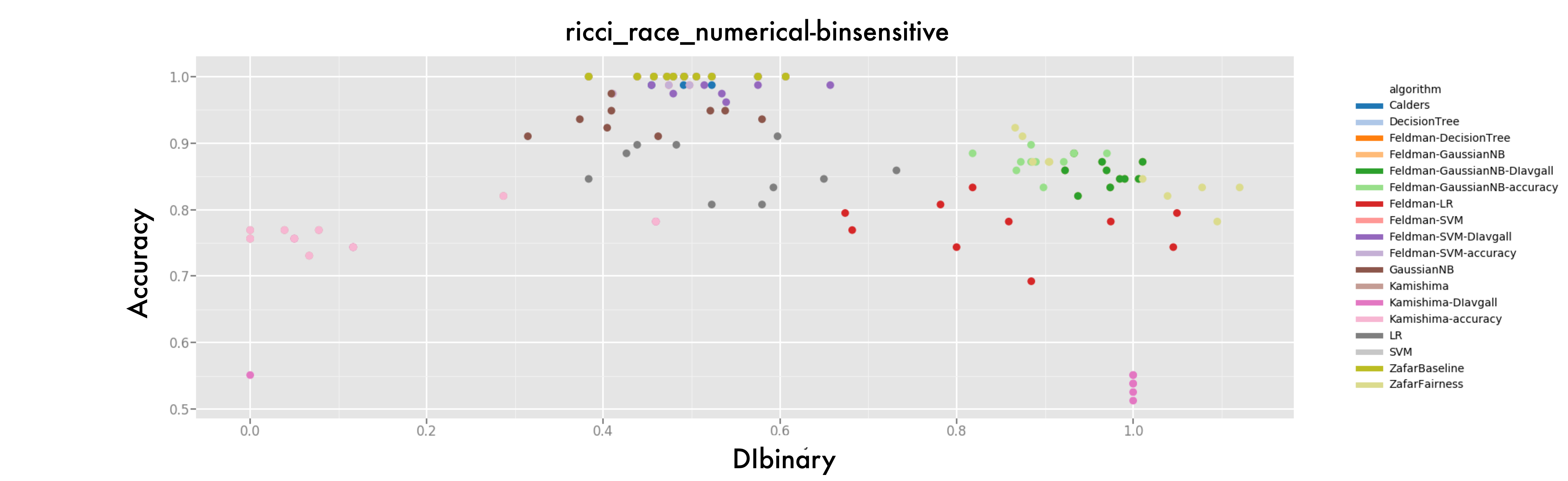}\\
  \includegraphics[width=\textwidth]{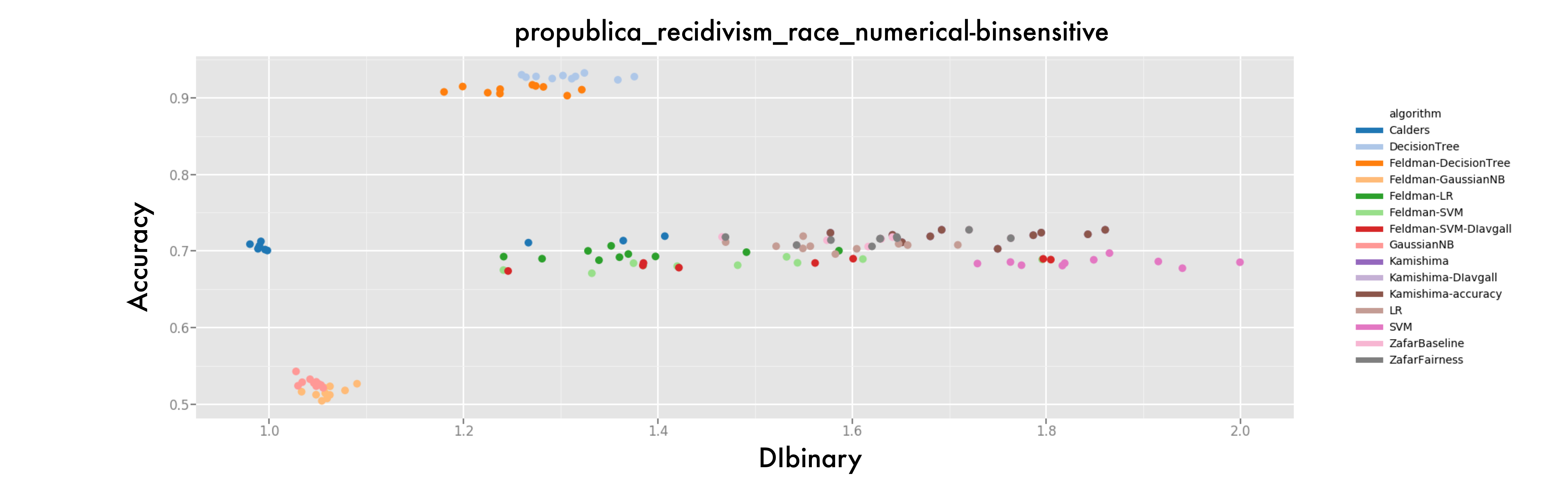}\\
  \includegraphics[width=\textwidth]{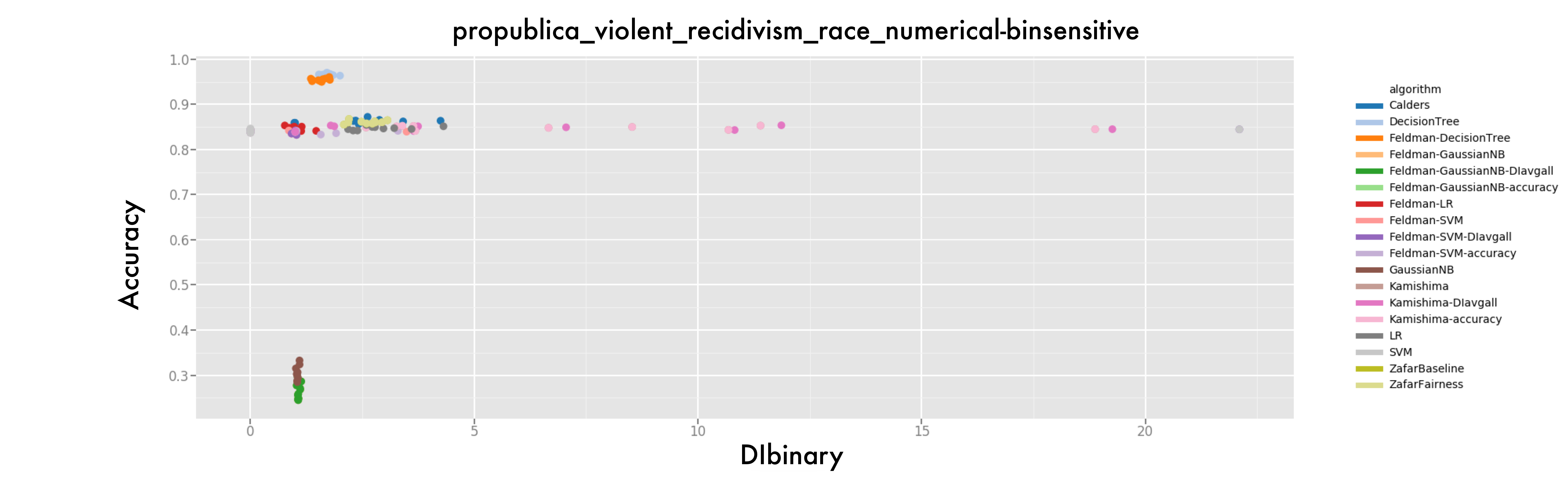}
\caption{The performance of all algorithms on each dataset with the goal of removing discrimination on a specific attribute.  From top to bottom, the algorithms and sensitive attributes considered are: Adult Income on race, German Credit on sex, Ricci on race, ProPublica recidivism on race, and ProPublica violent recidivism on race.  Each point is the result of a single algorithm running on a single training / test split - each algorithm is shown for ten such splits.}
\label{fig:alg_results}
\end{center}
\end{figure}

In Figure \ref{fig:alg_results} we can see a basic summary of the performance of each algorithm considered on each data set.  Since each algorithm focuses on creating a fair outcome with respect to a specific attribute in the data, we have chosen a single sensitive attribute to consider per dataset in these overall results.  It is clear that there is no one ``winner" - no algorithm that is both more fair and more accurate than the others on all datasets.  It is also clear that there is tremendous variation even within a single algorithm over the random splits it receives.  We examine this point in more detail next.

\subsection{Stability}
\label{sec:measuring-stability}
\begin{figure*}
  \centering
  \includegraphics[width=.45\linewidth]{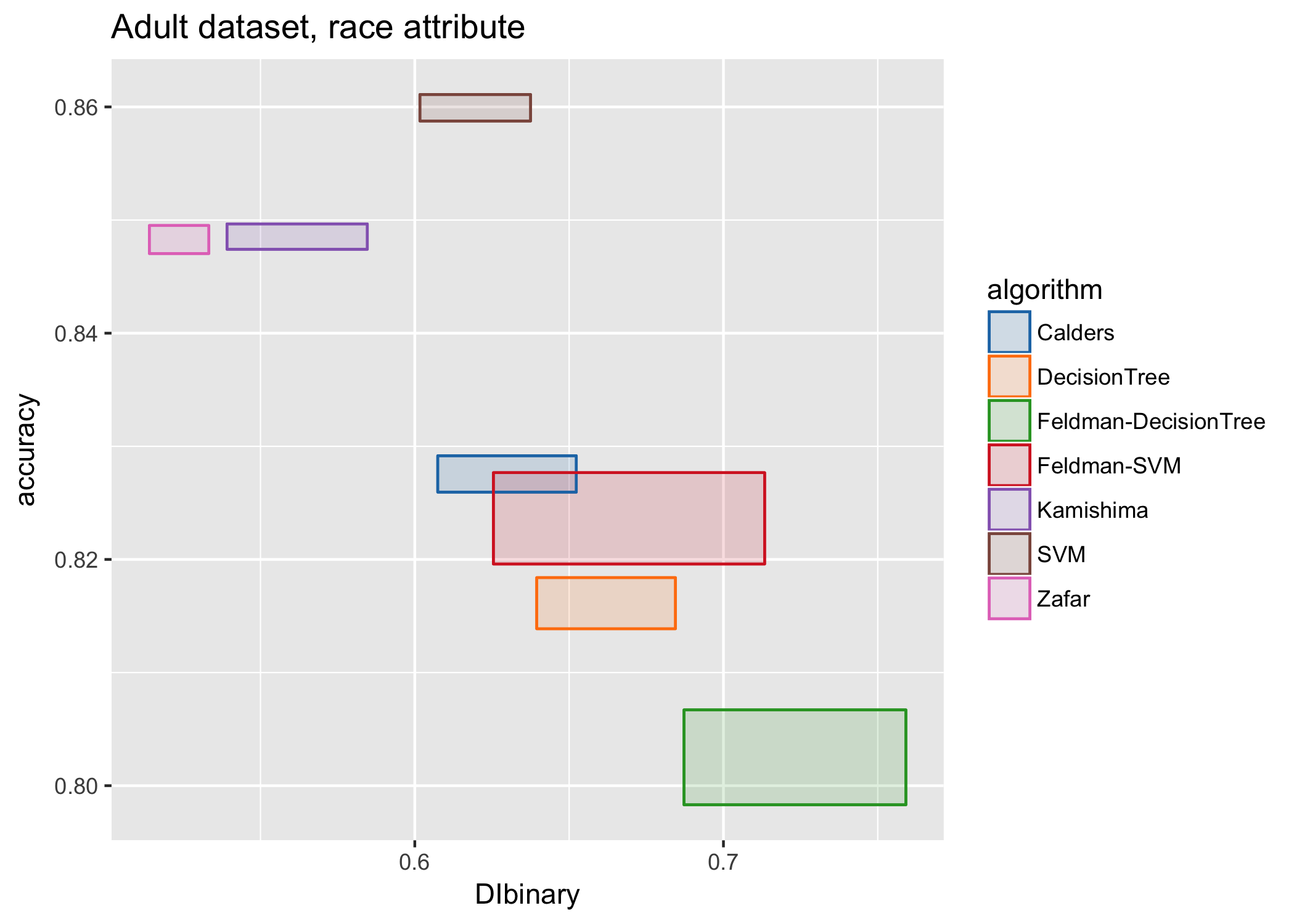}
  \includegraphics[width=.45\linewidth]{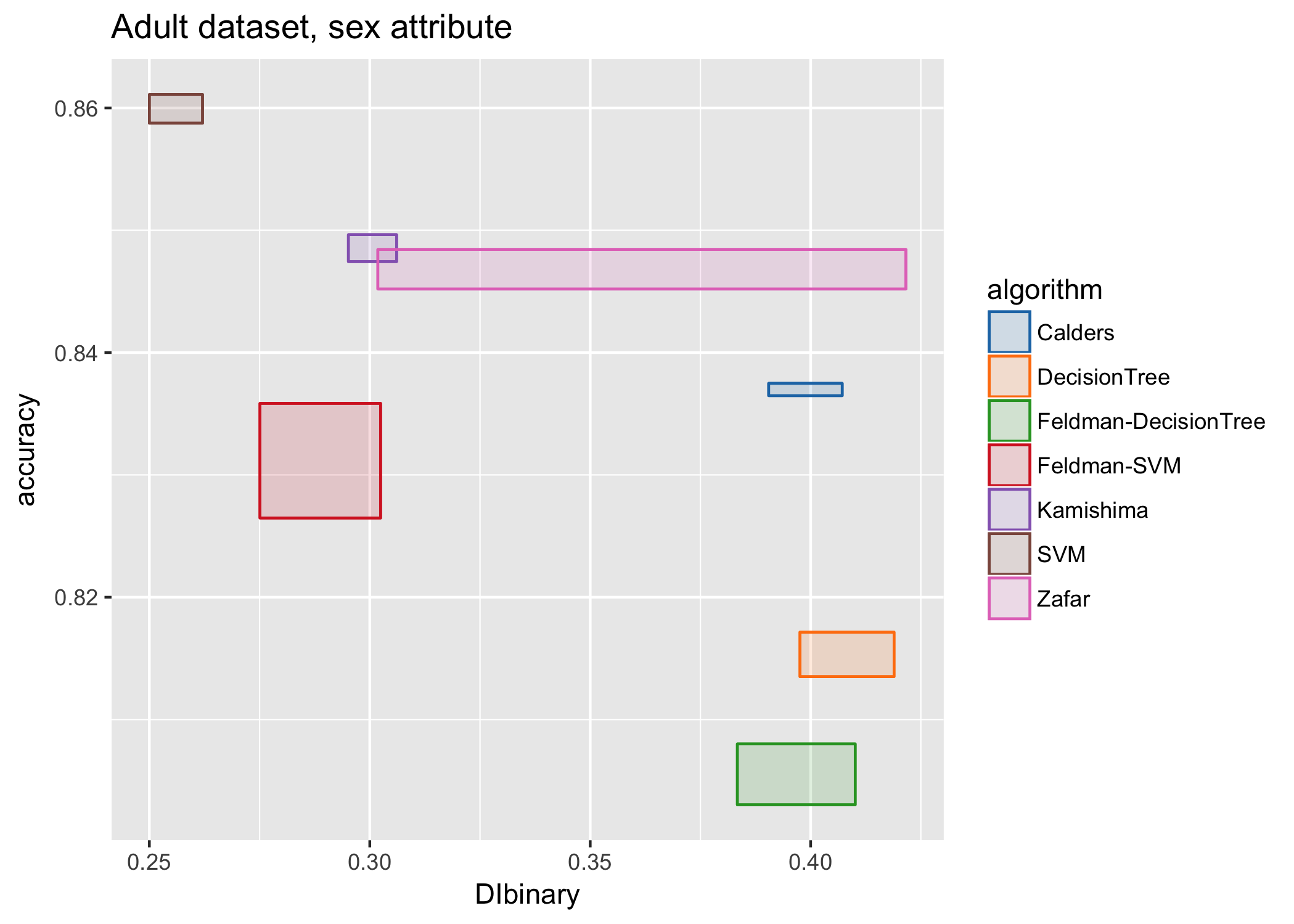}
  \caption{The stability of algorithms on the Adult dataset.  Each algorithm is tested on ten random train / test splits and a rectangle centered on the mean and with a width and height equal to the standard deviation along that measure is plotted.  On the left, the algorithms attempt to remove discrimination in terms of race, and on the right in terms of sex.
    }
  \label{fig:stability}
\end{figure*}

When analyzing algorithms, an additional question we are concerned with is that of \emph{stability} - will the algorithm still perform well if the training data is slightly different?  To assess this, we considered the standard deviation of each metric over 10 random splits.  The results are shown in Figure \ref{fig:stability} for the Adult Income data set for all algorithms when focusing on non-discrimination in terms of race (left) and sex (right) using \texttt{numerical+binary} preprocessing.  These results give perhaps the clearest indication of the quality of an algorithm on a given data set.  It is also easy to see that each algorithm occupies a slightly different place on the trade-off between fairness (measured here by DI when taken over binary sensitive attributes).  For example, when focusing on non-discrimination in terms of sex, the Zafar et al. algorithm is potentially the best choice in terms of a balance between fairness and accuracy, but the large standard deviation over DI may make it a less desirable option.  

\subsection{Parameters}
\label{sec:eval-tuning-param}
\begin{figure}
  \centering
  \includegraphics[width=.45\linewidth]{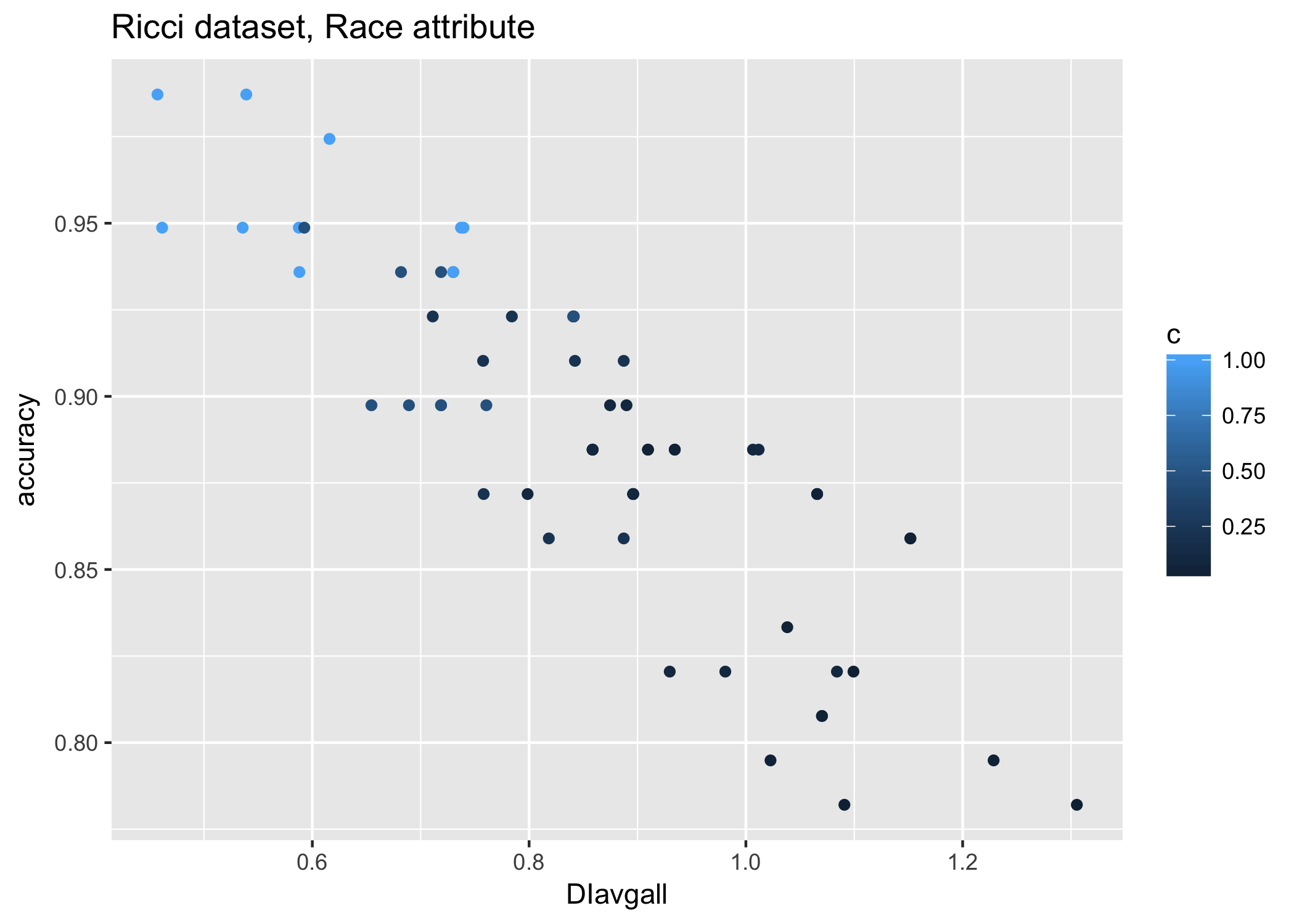}
  \includegraphics[width=.45\linewidth]{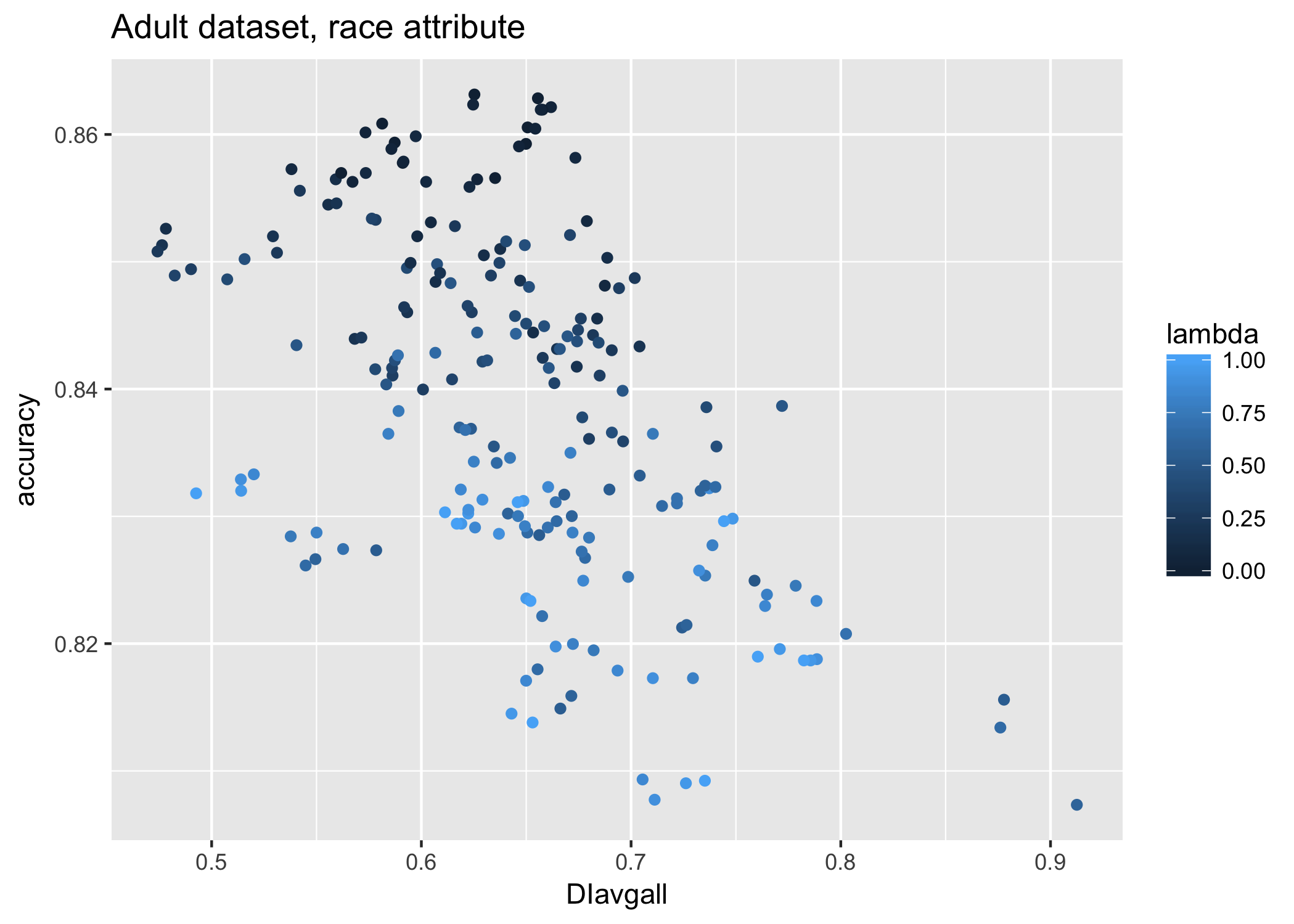}
  \caption{The results of the \citet{zafar2017fairness} algorithm on the Ricci dataset (left) and the \citet{2015_kdd_disparate_impact} algorithm on the Adult Income dataset (right) when the provided parameter to tradeoff between fairness and accuracy is used.  The parameter is varied and each split and each new parameter value is shown. }
  \label{fig:params}
\end{figure}

Many of these fairness-aware learning algorithms provide a parameter to allow a manual trade-off between fairness and accuracy.  We automate the search for this balance and present results for all algorithms optimizing accuracy or fairness.  This provides an additional means of testing the algorithm, as well as the possibility for further optimization of the tradeoff between the two.  In Figure \ref{fig:params} we show the different results based on parameter tuning for the \citet{zafar2017fairness} algorithm on the Ricci dataset (left) and the \citet{2015_kdd_disparate_impact} algorithm on the Adult Income dataset.
A clear tradeoff between fairness and accuracy in these algorithms can be seen; the parameters are appropriately allowing exploration of the possible solution space.

\subsection{Multiple sensitive attributes}
\label{sec:meas-handl-mult}

\begin{figure}[htb]
\begin{center}
\includegraphics[width=\textwidth]{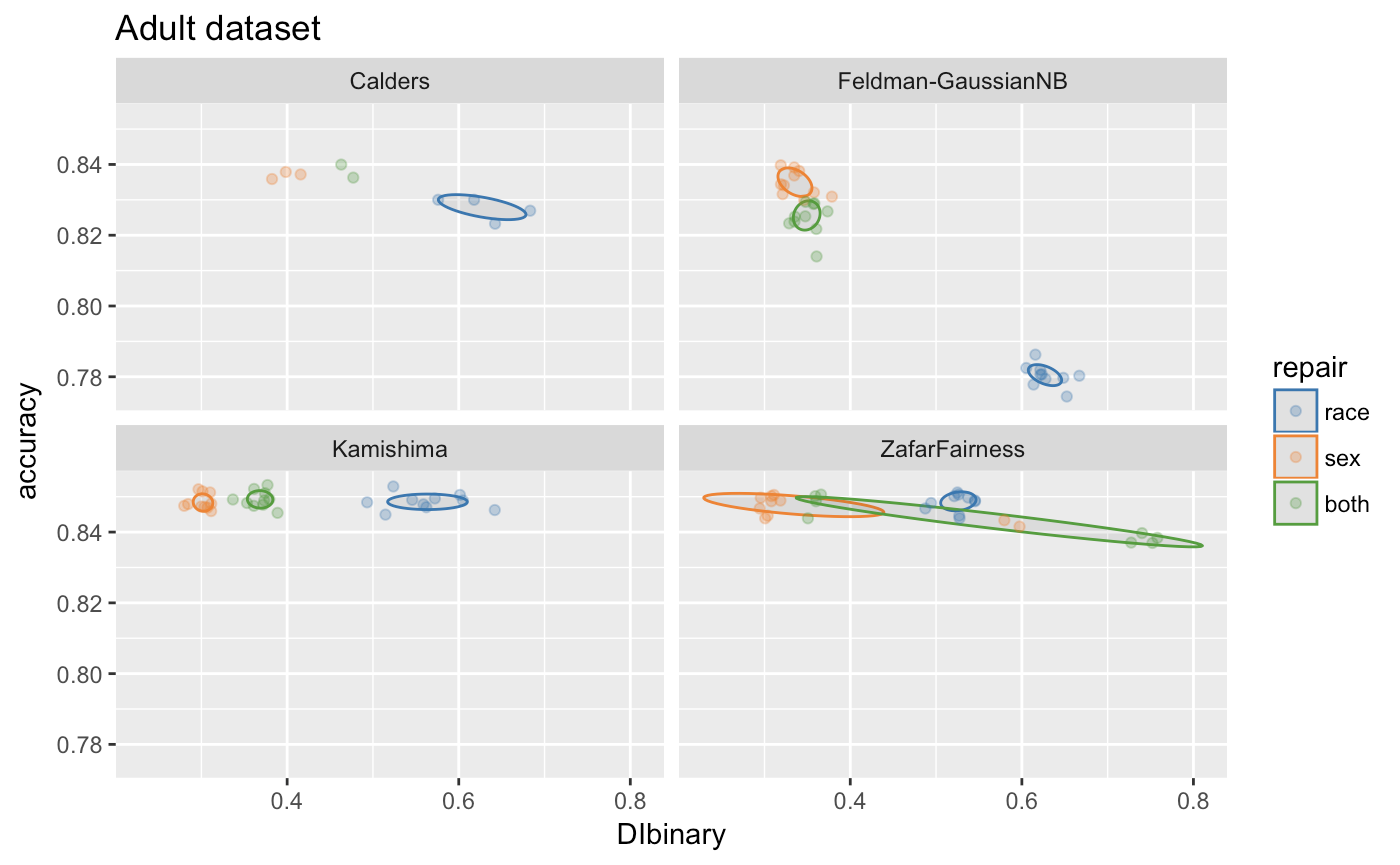}
\caption{Here, we show the behavior of four different algorithms when making predictions while accounting for different protected attributes (``repairing'' race and sex, as well as a composite attribute). Different algorithms not only behave quite differently from one another,  but their performance varies significantly depending on which specific attribute is being considered.}
\label{fig:multiple_sensitive}
\end{center}
\end{figure}

While there are still very few fairness-aware algorithms that can formally handle multiple sensitive attributes directly in the algorithm (\cite{kearns2017preventing,hebert-johnson_calibration_2017}), all algorithms discussed can handle them if preprocessed as described earlier so that they are combined into a single sensitive attribute (e.g., race-sex).  However, we might expect combining the attributes in this way to degrade performance under some metrics, especially in the case of algorithms that can only handle binary sensitive attributes, or when there are too many combinations for the size of the dataset to provide a large group of people with each new combined sensitive value.  Looking at the Adult dataset when fairness-aware algorithms are run focusing on non-discrimination in terms of race, sex, and both, we find varying results for each of the algorithms in Figure \ref{fig:multiple_sensitive}.  Sex is especially predictive on the Adult Income data set, so the DI value for sex is low, even on these fairness-aware algorithms.  Race generally receives a higher DI value from these algorithms.  When correcting for both at once, all of the algorithms find that the DI value is somewhere in between that for race and that for sex, but the \citet{zafar2017fairness} algorithm has a much larger variance over race and sex than over either individually.


\section{Discussion}
\label{sec:discussion--overall}

Besides providing a central point of access to existing
fairness-enhancing interventions and classification algorithms, our
benchmark also highlights a number of gaps in the current practice and
reporting of fairness issues in machine learning. We conclude with the
following recommendations for future contributions to the area:

\paragraph*{Emphasize preprocessing requirements.} If there are multiple plausible ways
in which a dataset can be processed to generate training data for an
algorithm, provide performance metrics for more than one of the
possible choices.  If algorithms are being compared to each other, ensure they are compared based on the same preprocessing.

\paragraph*{Avoid proliferation of measures.} A new measure for fairness
should only be introduced if it behaves fundamentally differently from
existing metrics. Our study indicates that a combination of
class-sensitive error rates and either DI or CV is a good
minimal working set.

\paragraph*{Account for training instability.}  Showing the performance
  of an algorithm in a single training-test split appears to be insufficient. We
  recommend reporting algorithm success and stability based on a moderate number of randomized training-test splits.

\bibliographystyle{ACM-Reference-Format}
\bibliography{fairness}

\end{document}